\documentclass[10pt,twocolumn,letterpaper]{article}

\usepackage[pagenumbers]{cvpr}    % To force page numbers, e.g. for an arXiv version

\usepackage[dvipsnames]{xcolor}
\usepackage{multirow}
\usepackage{xspace}

\newcommand{\proposal}{\textit{ImageNet-ES}\xspace}
\newcommand{\testbed}{\textit{ES-Studio}\xspace}

\definecolor{cvprblue}{rgb}{0.21,0.49,0.74}
\usepackage{natbib}
\usepackage{caption}
\usepackage{subcaption}
\usepackage{graphicx}
\usepackage{epsfig} 

\usepackage[pagebackref,colorlinks,citecolor=cvprblue]{hyperref}

\def\paperID{7880} % Paper ID 
\def\confName{CVPR}
\def\confYear{2024}

\title{Unexplored Faces of Robustness and Out-of-Distribution: \\Covariate Shifts in Environment and Sensor Domains
}

\author{Eunsu Baek \qquad Keondo Park \qquad Jiyoon Kim \qquad Hyung-Sin Kim\\
Seoul National University\\
\tt\small\{beshu9407, gundo0102, iamkjy, hyungkim\}@snu.ac.kr
}

\begin{document}
\maketitle
\normalsize 
\begin{abstract}
\vspace{-2ex}

Computer vision applications predict on digital images acquired by  a camera from physical scenes through light. However, conventional robustness benchmarks rely on perturbations in digitized images, diverging from distribution shifts occurring in the image acquisition process. 
To bridge this gap, we introduce a new %real-world 
distribution shift dataset, \proposal, comprising variations in environmental and camera sensor factors by directly capturing 202k images with a real camera in a controllable testbed.
With the new dataset, we evaluate out-of-distribution (OOD) detection and model robustness. We find that existing OOD detection methods do not cope with the covariate shifts in \proposal, implying that the definition and detection of OOD should be revisited to embrace real-world distribution shifts. We also observe that the model becomes more robust in both ImageNet-C and -ES by learning environment and sensor variations in addition to existing digital augmentations. Lastly, our results suggest that effective shift mitigation via camera sensor control can significantly improve performance without increasing model size.
With these findings, our benchmark may aid future research on robustness, OOD, and camera sensor control for computer vision. Our code and dataset are available at \href{https://github.com/Edw2n/ImageNet-ES}{https://github.com/Edw2n/ImageNet-ES}.
\end{abstract}
\vspace{-3ex}    
\section{Introduction}
\label{sec:intro}

The human vision system processes visual information by capturing light through the eyes and interpreting it within the brain.  While proper training of our brains is undoubtedly crucial, addressing eyesight or light-related challenges necessitates equipping ourselves with customized- or sun-glasses rather than relying solely on cognitive enhancement.

\begin{figure}[t]
\begin{center}
\begin{subfigure}{\linewidth}
    \includegraphics[width=\linewidth]{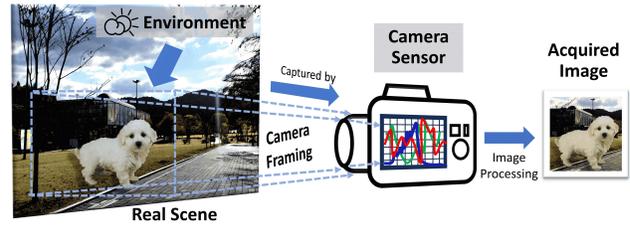}
    \vspace{-3ex}
    \caption{Real-world image acquisition process. Variations in the environmental and camera sensor factors can cause significant covariate shifts in the acquired image.}
    \label{fig:img-acquisition}
\end{subfigure}
\begin{subfigure}{\linewidth}
    \includegraphics[width=\linewidth]{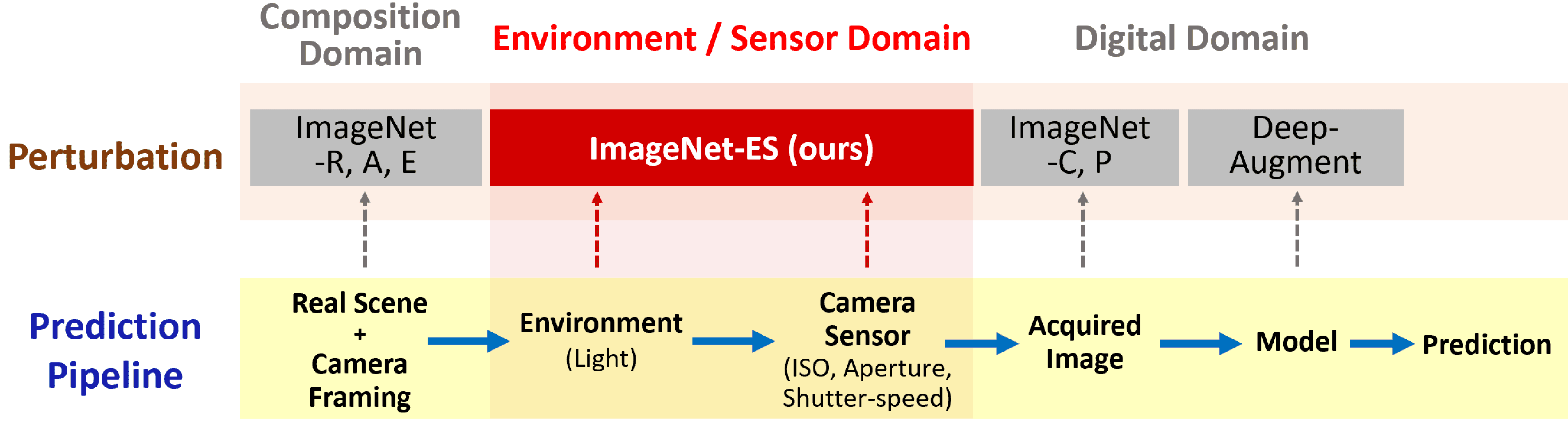}
    \vspace{-2ex}
    \caption{Real-world image prediction pipeline with existing perturbation benchmarks at each phase. \proposal first investigates the  environment and sensor domains directly, instead of mimicking via digital perturbation.}    
    \label{fig:pipeline}
    \vspace{-3ex}
\end{subfigure}
\end{center}
\caption{Motivation and contribution of \proposal, the first benchmark on the necessary but unexplored faces of image covariate shifts: environment and camera sensor domains.}
\vspace{-5ex}

\label{fig:pipeline_fig}
\end{figure}

Similarly, in many computer vision frameworks, as depicted in Figure~\ref{fig:img-acquisition}, a camera serves as the `eyes,' capturing authentic scenes through the play of light and generating digital images. These images are then interpreted by a deep neural network (\ie the brain), as illustrated in Figure~\ref{fig:pipeline}.
Continuous efforts towards improving AI systems to match the robustness of the human vision system predominantly focus on the `brain' component. Existing robustness benchmarks evaluate the resilience of model predictions against perturbations in \textit{digitized} images~\cite{hendrycks2019benchmarking,li2023imagenet,madry2018towards,Goodfellow2014ExplainingAH}. Various techniques, such as domain generalization/adaptation and out-of-distribution (OOD) detection, have refined deep learning models to handle distribution shifts~\cite{hendrycks2016baseline, liang2017enhancing, hendrycks2019scaling,
liu2020energy, react, lee2018simple, wang2022vim, sun2022out, ash, zhang2023openood, averly2023unified, hendrycks2021many,oquab2023dinov2}.

\begin{figure*}[t]
\begin{center}
\begin{minipage}{0.8\textwidth}
    \centering
    \vspace{-3ex}
    \includegraphics[height=8cm]{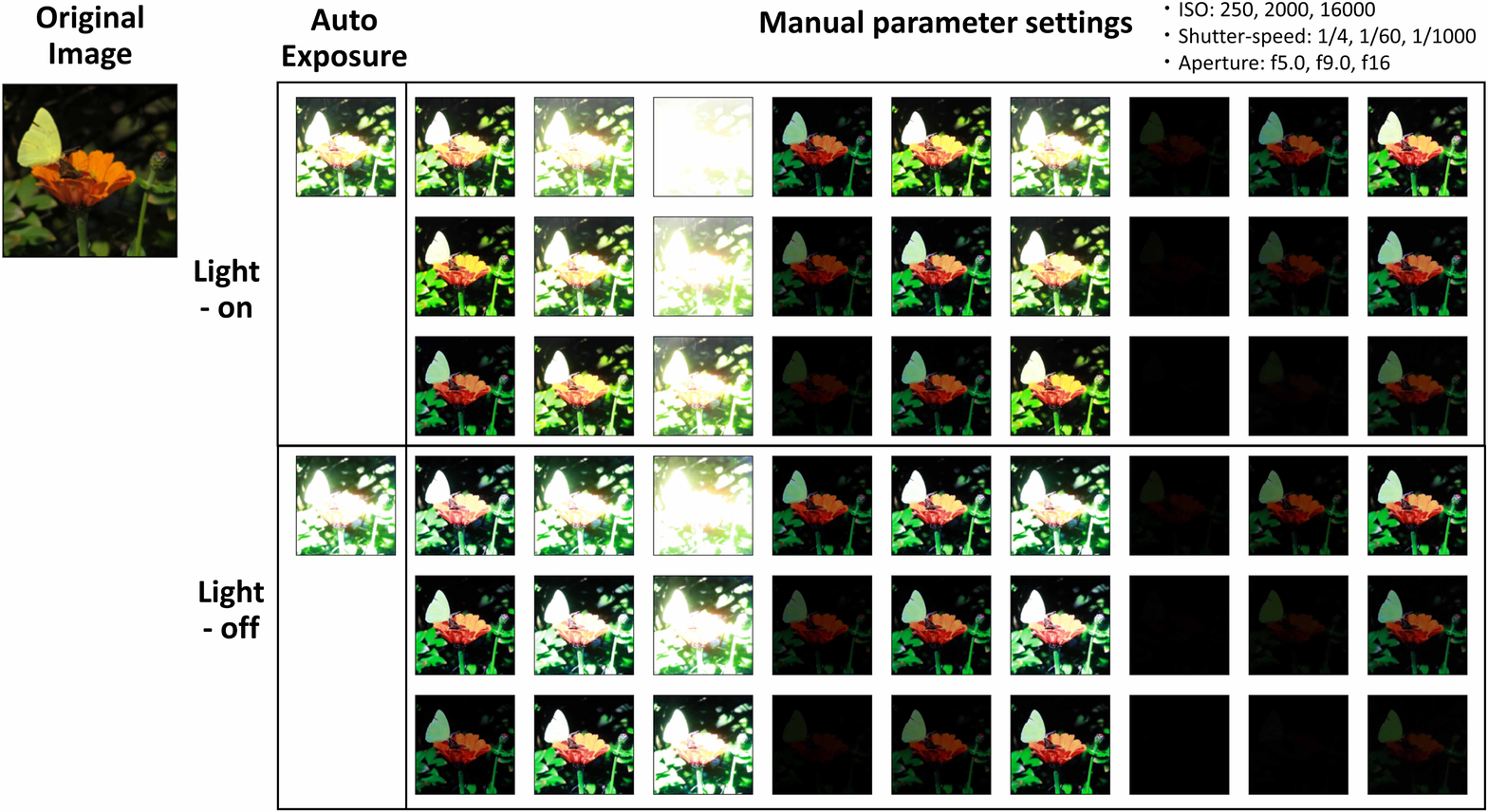}
\vspace{-4ex}
    \caption{Representative Examples of \proposal. In contrast to conventional robustness benchmarks that rely on digital perturbations, we directly capture 202k images by using a real camera in a controllable testbed. The dataset presents a wide range of covariate shifts caused by variations in light and camera sensor factors.}
    \label{fig:samples}
\end{minipage}
\hfill
\begin{minipage}{.18\textwidth}
    \centering
    \includegraphics[height=7.7cm]{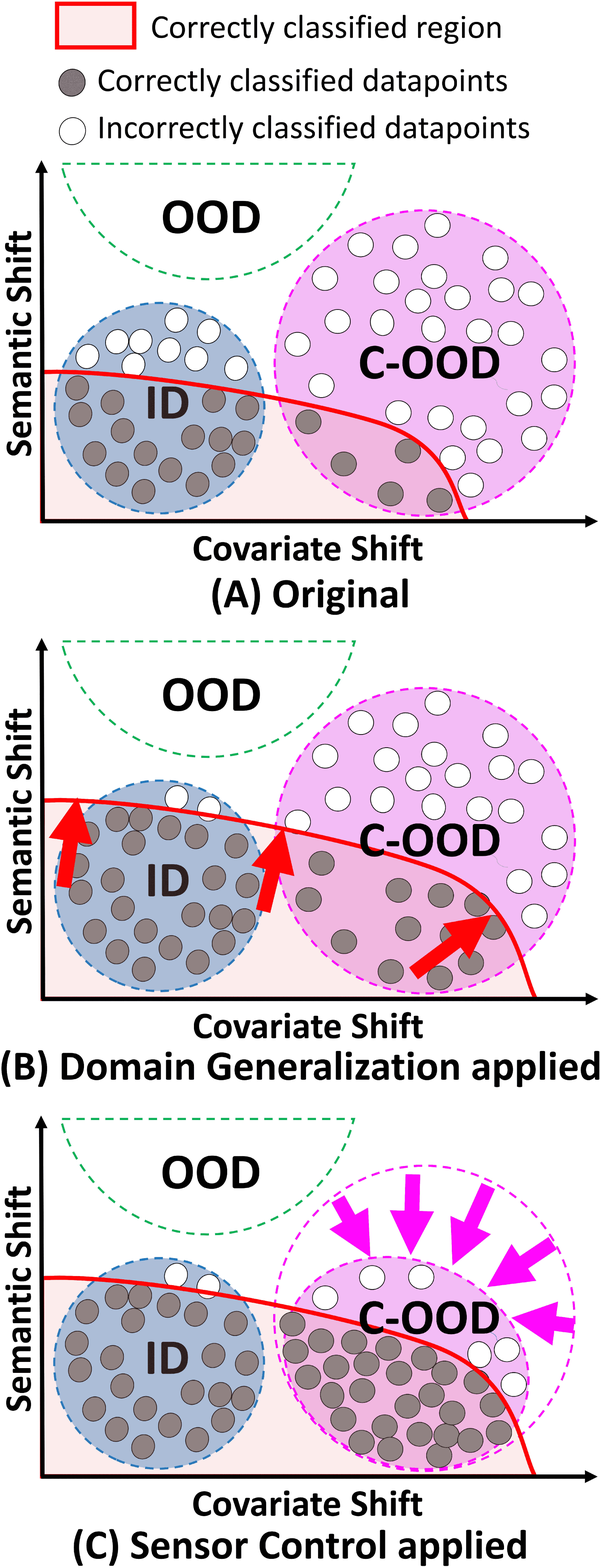}
    \vspace{-2ex}
    \caption{Robustness improvement scenario to cover real-world C-OOD}
    \label{fig:COOD}
\end{minipage}
\end{center}
\vspace{-5ex}
\end{figure*}

However, the implications of distribution shifts resulting from the \textit{image acquisition process} (\ie eyes), caused by variations in real-world light and camera sensor operations, remain unexplored. The absence of a benchmark %addressing this aspect 
introduces uncertainty regarding the generalizability of observed robustness in synthetic data to real-world applications. Moreover, the synergistic interplay between the camera sensor and the model has not been %thoroughly 
investigated. Therefore, current approaches may risk inefficiency by attempting to address eyesight/light problems through over-training the brain.

This work aims to narrow the gap between synthetic and real-world data by investigating the impact of environmental and camera sensor factors. Instead of relying on %indirect simulation through 
digital perturbation, we construct a controllable testbed, \testbed. This testbed allows us to directly capture images using a physical camera with varying sensor parameters (ISO, shutter speed and aperture) and different light conditions (on/off), resulting in a novel dataset called \proposal.

\proposal consists of 202k images covariate-shifted from 2,000 samples in TinyImageNet~\cite{wu2017tiny}. 
For example, Figure~\ref{fig:samples} shows 56 variations for a single sample, captured in \testbed under different light and camera sensor settings. 
These example images illustrate a broad spectrum of distribution shifts, suggesting that model robustness observed in conventional benchmarks might not necessarily generalize to our \proposal benchmark. 
Furthermore, some of the captured images even lose essential visual features due to severe perturbation, making them impractical  for model prediction. This implies that, as shown in Figure~\ref{fig:COOD}, restricting distribution shifts in the image acquisition phase via camera sensor control can be more practical than solely focusing on model improvement.

With the \proposal dataset, we conduct an extensive empirical study on OOD detection and domain generalization. Furthermore, we explore the potential of camera sensor control in addressing real-world distribution shifts. Our study unveils a series of noteworthy findings as follows:
\begin{itemize}
    \item \textbf{OOD definition:} Covariate-shifted data (C-OOD) have been categorized entirely as either OOD or in-distribution (ID). However, C-OOD data in \proposal exhibit widespread OOD scores in most metrics, including both ID and OOD. 
    Model-Specific OOD (MS-OOD)~\cite{averly2023unified} is more proper for fine-grained labeling of our C-OOD data.
    \item \textbf{OOD detection:} State-of-the-art (SOTA) OOD detection methods, focusing on distinguishing semantic shifts, falter in \proposal. OOD detection should be improved to incorporate real-world covariate shifts together.
    \item \textbf{Domain generalization:} Existing digital augmentations do not incorporate distribution shifts in \proposal. Learning  environment/sensor-domain perturbations in \proposal  with existing augmentations improves model robustness, even in conventional benchmarks.
    \item \textbf{Potential of sensor control:} Camera sensor control can significantly improve prediction accuracy by mitigating distribution shifts. With sensor control, EfficientNet can perform comparably to much heavier transformer models.
    \item \textbf{Direction of sensor control:} High-quality images in terms of model prediction do not necessarily align with human aesthetics but rather with what the model learns from training data. Sensor control should be grounded in the features that the model (not the human) prefers. 
\end{itemize}
Overall, %our results show that 
future research on OOD detection and model robustness requires more thorough evaluations, including environmental and camera sensor variations. Furthermore, it is valuable %for future research 
to explore camera sensor control so that acquired images contain more features preferred by the model.

\section{Related Work}

%%%%%%%%%%%%%%%%%%%%%%%%%%%%%%%%%%%%%%%%%%%%%%%%%%%%%%%
%%%%%%%%%%%%%%%%%%%%%%%%%%%%%%%%%%%%%%%%%%%%%%%%%%%%%%%
\subsection{Robustness Benchmarks}

A number of benchmarks have employed various digital perturbations to assess image classifier robustness or OOD detection methods.
Notably, ImageNet-C and -P~\cite{hendrycks2019benchmarking} simulate environmental and adversarial perturbations through blur, noise, brightness, etc.
ImageNet-A and -O~\cite{hendrycks2021natural} limit spurious cues using adversarial perturbations. 
Several datasets utilize visual renditions to change real scenes, such as art, cartoons, patterns, toys, paintings, etc. ~\cite{hendrycks2021many, wang2019learning, rusak2022imagenet}. 
SVSF~\cite{hendrycks2021many} or ImageNet-E~\cite{li2023imagenet} changes camera views or image compositions.

While these benchmarks aim to incorporate real-world distribution shifts, such as camera framing, their approaches are limited to the indirect simulation of actual shifts via perturbing already-acquired digital images. 
Recent studies have highlighted that SOTA OOD detection methods face challenges due to a lack of knowledge about the real-world OOD distributions~\cite{watermarking2022} and experience performance degradation in near-OOD, shifted benchmarks~\cite{raisingbar2023}. 
Building on prior work, our \proposal dataset directly modifies physical light and camera sensor parameters, which provide another type of real-world distribution shifts and demystify the relationship between digital and physical manipulations.

%%%%%%%%%%%%%%%%%%%%%%%%%%%%%%%%%%%%%%%%%%%%%%%%%%%%%%%
%%%%%%%%%%%%%%%%%%%%%%%%%%%%%%%%%%%%%%%%%%%%%%%%%%%%%%%
\subsection{Out-of-Distribution (OOD) Detection}

Out-of-distribution (OOD) detection is the task of identifying test data that come from a distribution  different from the distribution of training data, due to either semantic shift (S-OOD) or covariate shift (C-OOD)~\cite{raisingbar2023}.

OOD studies have focused on detecting samples with semantic shifts (S-OOD) that do not belong to any of the classes present in the training set. 
A number of methods determine the OOD score based on the decision-making component of classifiers~\cite{hendrycks2016baseline,liang2017enhancing, hendrycks2019scaling, liu2020energy}. These techniques are more robust when class-agnostic information needs to be carefully considered, but  vulnerable to significant semantic shifts or overconfidence issues~\cite{react, wang2022vim}. 
To alleviate these problems, other methods calculate the OOD score based on features the model learned~\cite{react, lee2018simple, wang2022vim, sun2022out, ash}. Rigorous efforts in this area have achieved nearly perfect performance.

However, prior work has relatively unexplored how to handle covariate-shifted (C-OOD) samples. 
A handful of studies have considered entire C-OOD examples as in-distribution (ID) to enhance classifier robustness against covariate shifts~\cite{yang2021semantically, yang2023full, zhang2023openood}. Some studies have taken opposite approaches, treating all C-OOD samples as OOD to make OOD detection more generalizable to non-semantic shifts~\cite{hsu2020generalized}. 
To address the problem of the rough treatment of entire covariate-shifted data as ID or OOD, more recent studies provide fine-grained categorization of C-OOD samples into ID and OOD, based on their own definitions~\cite{tian2021exploring, averly2023unified}. 
Notably, Averly and Chao have proposed a unified criterion that incorporates both S-OOD and C-OOD data based on model prediction results, called Model-Specific OOD (MS-OOD)~\cite{averly2023unified}. MS-OOD reveals the problems of existing methods but does not provide a solution.

Looking forward, OOD detection should be improved to reliably handle both S-OOD and C-OOD data with a well-defined OOD score and detection method, which requires support from proper benchmarks. \proposal can contribute to this aspect by providing realistic C-OOD samples.

%%%%%%%%%%%%%%%%%%%%%%%%%%%%%%%%%%%%%%%%
%%%%%%%%%%%%%%%%%%%%%%%%%%%%%%%%%%%%%%%%
\subsection{Domain Generalization}

Domain generalization focuses on improving the robustness of models to distribution shifts in testing domains.  
To this end, Hendrycks \etal identified that using larger models and artificial data augmentations (called DeepAugment) can improve model robustness~\cite{hendrycks2021many}. While many augmentation techniques~\cite{Cubuk, hendrycks2020augmix, yun2019cutmix} have shown to improve the robustness, their evaluation scope is limited to digital corruptions.
More recently, foundation models have demonstrated success in learning effective feature representations through architectural changes~\cite{liu2022convnet}, discriminative self-supervised pretraining~\cite{he2022masked, zhou2021ibot, oquab2023dinov2}, or large uncurated data~\cite{oquab2023dinov2}.

However, prior work %on domain generalization 
has focused on digital distribution shifts (\eg pixelate or gaussian noise etc.), scene and camera composition shifts. On the other hand, our \proposal addresses other types of distributional shifts, such as those arising from the image acquisition process.

\vspace{-1ex}

\section{Background}

 As illustrated in \cref{fig:pipeline_fig}, an image is influenced by three primary aspects at the point of its capture. Firstly, the term \textit{composition} pertains to the arrangement, organization, and layout of visual elements within the frame of the image. Composition is subject to dynamic alterations caused by the movement of objects, addition or removal of objects, or other modifications. Camera operations, such as zooming or tilting, can also impact  the resulting image. 
 Secondly, \textit{environment} signifies lighting conditions. For example, light can be scattered by dust or smoke, leading to image blurring. The position and intensity of light can also affect the image's quality. 
 Finally, \textit{camera sensor} generates an image from the light. The captured image can dynamically fluctuate according to sensor parameter settings. 
 
 This work focuses on variations in environmental and sensor factors without changing the composition.

\subsection{Camera Operation for Image Acquisition}

Before digitization, image variations can be introduced during the camera's  acquisition process, which involves the following steps: (1) light reception, (2) sensor conversion, (3) image signal processing, and (4) final image creation.

Firstly, light is captured from the scene and environment. This light, the primary source of image variation in photography, plays a crucial role in determining the quality and characteristics of the image.
Next, the captured light hits the camera sensor, which converts the light into an electrical signal. The types and settings of the sensor can influence the image, with different sensors responding differently to the same light conditions. 
The electrical signal undergoes processing by the camera's internal systems. This processing commonly includes operations such as noise reduction, white balance adjustment, and color grading. Finally, the processed signal is converted into an image.

While image signal processing techniques can introduce various perturbations and contribute to the quality improvement of the final image, these results are \textit{fundamentally bounded} by the original electrical signal generated by the sensor from the lighting conditions. Therefore, despite numerous existing perturbations through post-processing, investigating the impact of environmental and sensor factors has additional value.

%%%%%%%%%%%%%%%%%%%%%%%%%%%%%%%%%%
%%%%%%%%%%%%%%%%%%%%%%%%%%%%%%%%%%
\subsection{Light Factor in Environment Domain}

In the environmental domain, changes in lighting conditions significantly impact the captured image.  
For example, an object photographed under bright overhead lighting may cast a strong shadow, altering the object's appearance. Similarly, an object photographed in low light may lack sufficient detail. Furthermore, changes in the color of the light, such as transitioning from daylight to artificial light, can affect how colors appear in the image. 
These variations present challenges for deep learning models, which often rely on consistent lighting for accurate image recognition.

\begin{table*}[t]
\centering
  \caption{Environment and Sensor specifics of \proposal collection}
  \vspace{-2ex}
  \label{tab:collection}

\resizebox{\linewidth}{!}{ %< auto-adjusts font size to fill line
  \begin{tabular}{ccccccccccc}
    \toprule
    Dataset       & Original samples & Light                    & Camera sensor    & ISO & Shutter speed & Aperture &  Captured images \\
    \midrule
    \multirow{2}{*}{Validation}   & 1,000   & \multirow{2}{*}{On/Off} & Auto exposure (5 shots)        & Auto  & Auto  & Auto  &  10,000 \\
                           & (5 samples/class) &                               & Manual (64 options)        & 200/800/3200/12800 & (0"4')/(1/20')/(1/160')/(1/1250') & f5.0/f9.0/f13/f20      & 128,000 \\
    \hline
    \multirow{2}{*}{Test}  & 1,000                                 & \multirow{2}{*}{On/Off} & Auto exposure (5 shots)        & Auto  & Auto  & Auto  & 10,000 \\
                           &   (5 samples/class)  &                               & Manual (27 options)        & 250/2000/16000 & (1/4')/(1/60')/(1/1000') & f5.0/f9.0/f16   & 54,000 \\
  \bottomrule
    \end{tabular}
} % \resizebox
\vspace{-3ex}
\end{table*}

%%%%%%%%%%%%%%%%%%%%%%%%%%%%%%%%%
%%%%%%%%%%%%%%%%%%%%%%%%%%%%%%%%%
\subsection{Light Factor in Camera Sensor Domain}

The camera sensor has three main parameters, ISO, shutter speed, and aperture, which influence light levels in an image while also impacting various aspects of the captured scene.

\begin{itemize}
    \item \textbf{ISO} adjusts the sensitivity of the camera sensor to light. Higher ISO values increase brightness but may introduce additional noise to the image. 
    
    \item \textbf{Shutter speed} governs the duration that the camera's shutter remains open. Slower shutter speed allows more light to reach the sensor, resulting in a brighter image but also motion blur. Conversely, faster shutter speed can produce a darker image and freeze motion. 

    \item \textbf{Aperture} determines the size of the lens opening, regulating light entry and affecting the image's depth of field. A larger aperture brightens the image but leads to a shallower depth of field, concentrating focus on a limited portion of the scene.
\end{itemize}

While manual control of these parameters is possible, most cameras are equipped with automatic exposure control. The \textbf{auto exposure} function calculates the optimal exposure settings for a given scene. However, it is important to note that the optimal settings are for human aesthetic, which may not align with those optimal for model predictions. 
\section{\testbed and \proposal}
\vspace{-0.5ex}

To compensate the missing perturbations in current datasets, we construct a new testbed, \textbf{\testbed} (\textbf{E}nvironment and camera \textbf{S}ensor perturbation \textbf{Studio}). It can control physical light and camera sensor parameters during data collection. Utilizing \testbed, we compile \textbf{\proposal}, a novel dataset comprising 202,000 samples of perturbed data from the environment and camera sensor domains.

\subsection{\testbed Design Considerations and Setup}

\begin{figure}
\begin{center}
    \includegraphics[width=1.0\linewidth]{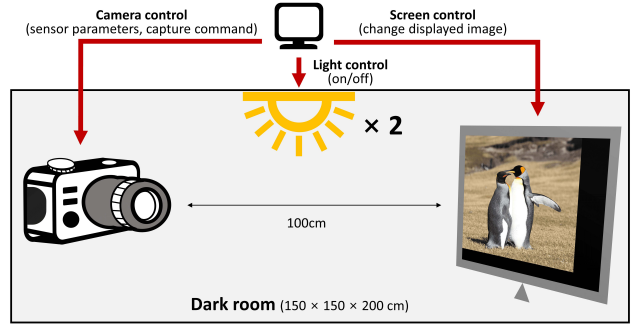}
\end{center}
\vspace{-4ex}
\caption{Illustration of the \testbed setup}
\vspace{-3ex}
\label{fig:testbed_fig}
%\vspace{-2ex}
\end{figure}

In constructing our data collection studio, we prioritize two main considerations: 1) ensuring reproducibility and 2) capturing real-world perturbations, particularly those related to light factors in both the environment and camera sensor domains. 
Specifically, being the first effort to gather such real-world perturbations, it is crucial for our data collection process to be reproducible, facilitating and promoting future research in this area. To achieve this, we have employed \testbed, providing individual control over environment and sensor parameters involved in image acquisition. 

The construction of \testbed is depicted in Figure~\ref{fig:testbed_fig}. First, we established  a completely dark room to eliminate any external light during the data collection process. 
The room is equipped with four main components: (1) a large screen to display the reference dataset, (2) a camera with adjustable parameters for ISO, shutter speed, and aperture, (3) two ceiling lamps to manipulate environmental light, and (4) a desktop and Wi-Fi network to manage above components. 
More details of \testbed settings are in Appendix. %\section{Appendix}

%%%%%%%%%%%%%%%%%%%%%%%%%%%%%%%%%%%%%%%%%%%%%%%%%%%
%%%%%%%%%%%%%%%%%%%%%%%%%%%%%%%%%%%%%%%%%%%%%%%%%%%
%\vspace{-1ex}
\subsection{\proposal dataset}
\subsubsection{Sampling Process for Target Datasets}

We choose Tiny-ImageNet~\cite{wu2017tiny}, a 200-class subset derived from ImageNet-1K, as our reference dataset. This dataset not only provides a diverse range of categories but also demands less computational power for experiments compared to ImageNet-1K. 
We randomly select ten images from each category in the validation set of Tiny-ImageNet. Subsequently, we divide these images into two halves, utilizing the first five for validation and the remaining five for testing purposes. To ensure visual fidelity, each sampled image maintains a resolution greater than $375 \times 500$ pixels, preventing distortion when displayed on the screen. In total, we systematically sample 2,000 images. 

\vspace{-2ex}
\subsubsection{Data Collection} 

Table~\ref{tab:collection} provides a comprehensive overview of the collected data. We display each sampled reference image on the screen and take its picture  multiple times while varying the environmental and camera sensor factors.  

We consider two options for the environmental factor: lights in the ``on'' and ``off'' states.
For camera sensor control, we use both auto exposure and manual parameter settings. 
Under auto exposure, the camera autonomously determines each sensor parameter. Given that the auto-controlled parameters can be different at each time, we capture each sample five times to observe the average effect.
For manual parameter setting, we use four different options for ISO, shutter and aperture during the validation split, and three options during the test split, leading to 64/27 variations in the validation/test split. To ensure the integrity of our data collection process, we implemented pauses between setting changes. Specifically, we introduced a one/seven/ten-second pause between each parameter option, between changes in light options, and between sample image changes, respectively. A detailed log is recorded for each image, serving debugging purposes.

\vspace{-2ex}

\subsubsection{Data Processing and Validation}

The next step involves cropping the valid image area from the collected images. 
The valid image area is determined through a systematic process: First, we display a visually discriminative reference image on the screen and capture the screen with the camera. We extract crucial information for the  captured image, including the left top point, width and height of the reference image. 
Then, for other images, we determine the valid area of each image by using the digitally calculated ratio of each image to the reference image. Finally, we set the padding to the determined valid area and crop the captured image accordingly.

To validate %the correctness of 
the \proposal collection, we conduct a subjective validation approach. For each reference sample, we aggregate all images taken under different settings and concatenate them into a single image along with the original sample. This composite image is then reviewed by three individuals to ensure that all images are captured consistently. 
The validation process also confirms that the collected images align accurately with the original image.

\section{Experiments}

We design experiments to evaluate the impact of distribution shifts within the environmental and camera sensor domains. The experiments include widely used methods for OOD detection and domain generalization.

%%%%%%%%%%%%%%%%%%%%%%%%%%%%%%%%%%%%
%%%%%%%%%%%%%%%%%%%%%%%%%%%%%%%%%%%%
\subsection{OOD Detection}
%\subsubsection{Setup}

We validate OOD detection techniques on \proposal: ViM~\cite{wang2022vim}, ReAct~\cite{react}, ASH~\cite{ash}, MSP~\cite{hendrycks2016baseline} and ODIN~\cite{liang2017enhancing}. They report SOTA performance and serve as baseline methods in recent OOD studies~\cite{zhang2023openood, averly2023unified}. Likewise, EfficientNet-B0~\cite{pmlr-v97-tan19a} is selected as the underlying model for OOD detection, given its widespread use in OOD studies. Training details and evaluations for other models are in Appendix.

\vspace{-2ex}
\subsubsection{Evaluation of OOD Definition}

\begin{figure}[t]
    \centering
        \begin{subfigure}{0.48\linewidth}
            \includegraphics[width=\linewidth]{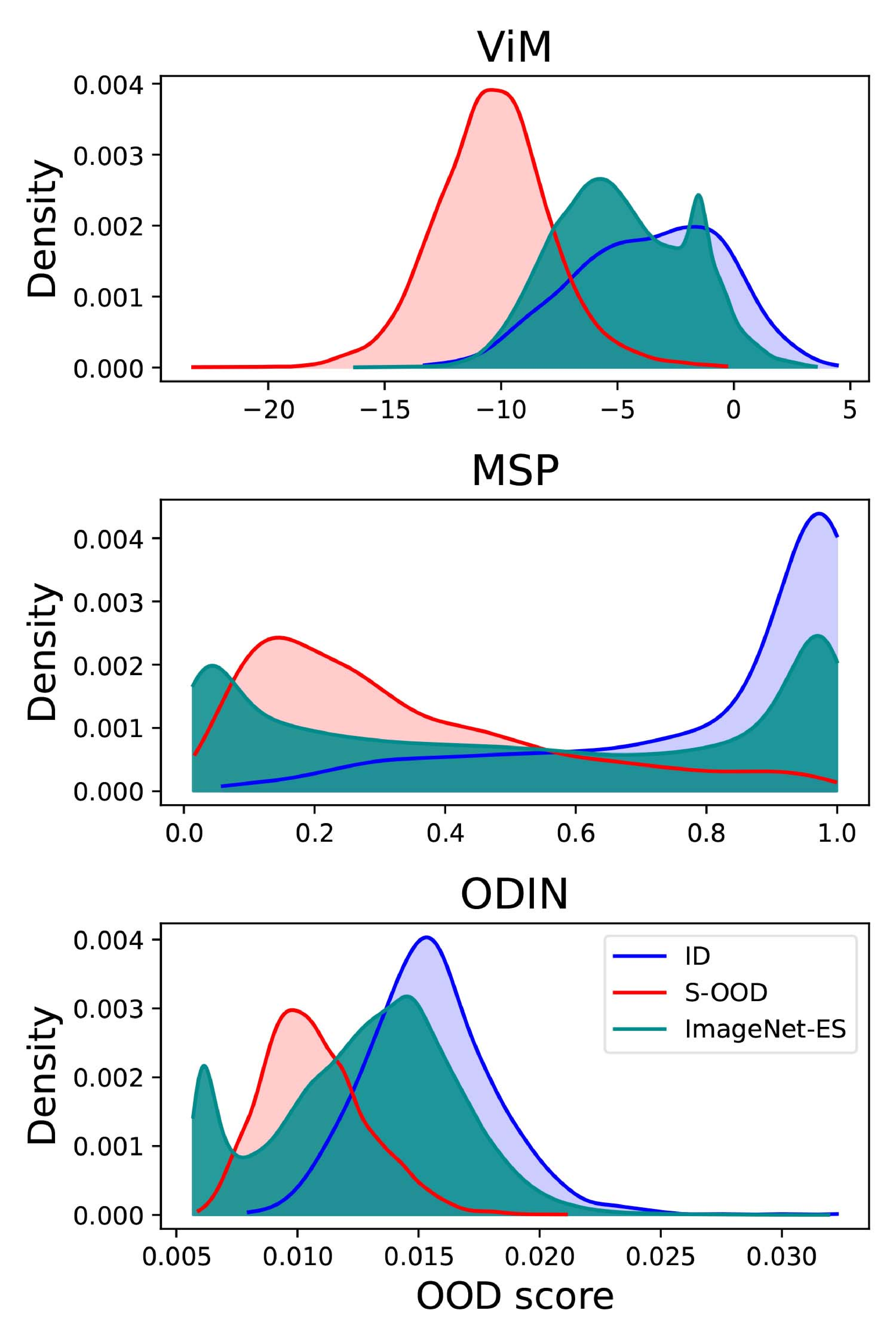}
            \vspace{-4ex}
            \caption{Semantics-centric framework}
            \label{fig:ss-dist}
        \end{subfigure}
        % \hspace{0.5cm}
        \begin{subfigure}{0.48\linewidth}
            \includegraphics[width=\linewidth]{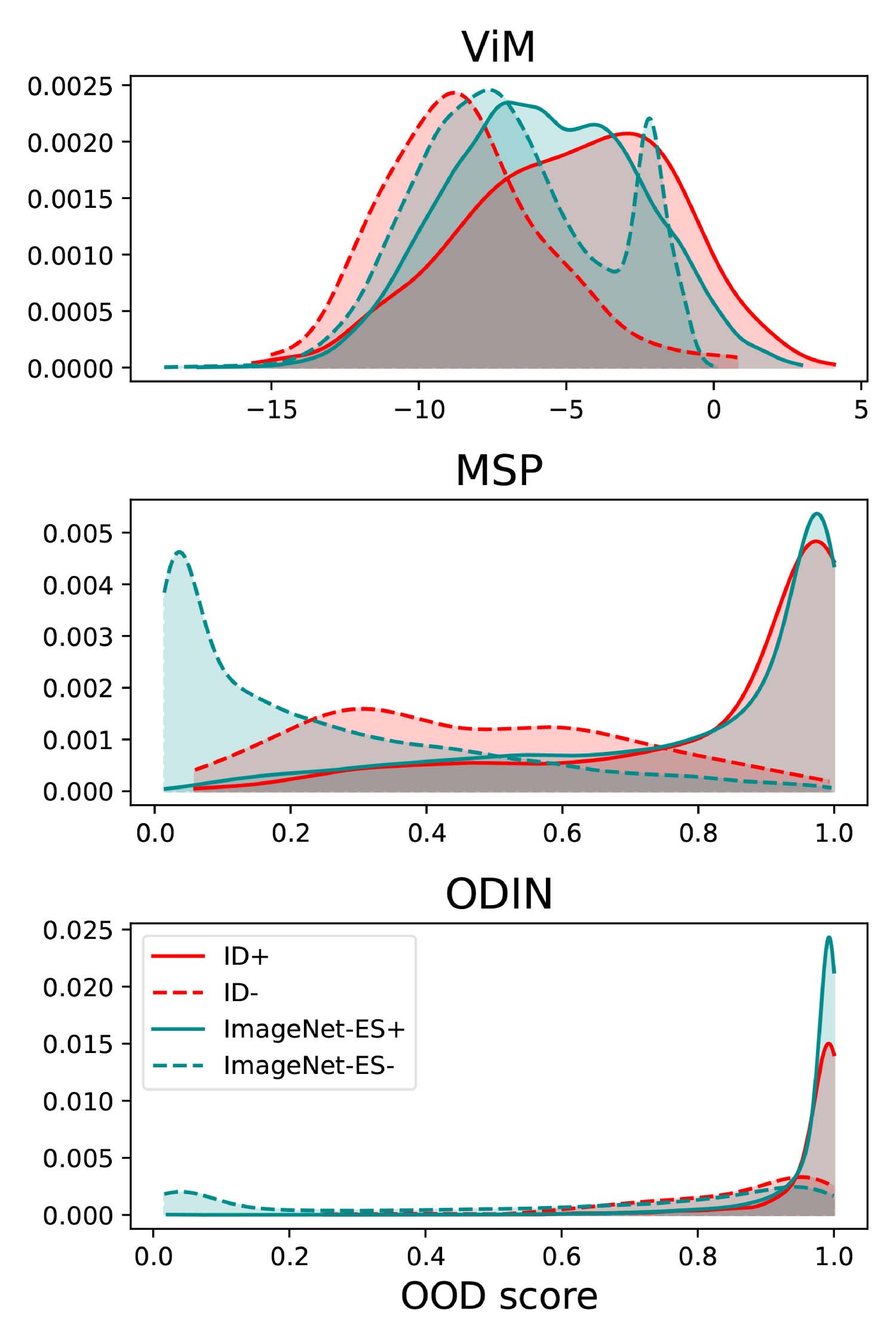}
            \vspace{-4ex}
            \caption{MS-OOD framework}
            \label{fig:MS-OOD-setting}
        \end{subfigure}        
    \vspace{-1ex}
    \caption{OOD score distribution with semantics-focused and MS-OOD frameworks. Tiny-ImageNet~\cite{wu2017tiny} and Texture~\cite{Cimpoi2014CVPR} are used for the ID and S-OOD datasets, respectively. \proposal serves as a C-OOD dataset.}
    \vspace{-3ex}
    \label{fig:conventional-vs-MS-OOD}
\end{figure}

Most OOD detection techniques are developed under a  framework that focuses on detecting samples with semantic shifts (S-OOD). Under this framework, all samples from \proposal (\ie C-OOD data) should be classified as either OOD or In-Distribution (ID) in their entirety. 
To assess the validity of the semantics-centric OOD definition under \proposal, we analyze the distribution of OOD scores of ViM~\cite{wang2022vim}, MSP~\cite{hendrycks2016baseline} and ODIN~\cite{liang2017enhancing} on ID (Tiny-ImageNet~\cite{wu2017tiny}), S-OOD (Texture-O~\cite{Cimpoi2014CVPR}) and \proposal (C-OOD) datasets, as in Figure~\ref{fig:ss-dist}. 
While OOD detection techniques provide clearly distinguished OOD scores for ID (blue region) and S-OOD (red region), the scores on \proposal (green region) are widely spread across the entire spectrum between OOD and ID. The results show that treating entire C-OOD data in \proposal as either ID or OOD leads to significant detection errors. It is risky to directly apply the semantics-centric framework in the presence of C-OOD data. 

We also evaluate an alternative framework, called MS-OOD (Model-Specific OOD)~\cite{averly2023unified}. In this framework, OOD is defined by considering model-specific acceptance (MS-A) or rejection (MS-R): 
(1) MS-A includes ID and C-OOD samples that are correctly classified by the model, denoted as ID$+$ and C-OOD$+$. (2) On the other hand, MS-R includes all S-OOD samples, as well as ID and C-OOD samples that are misclassified by the model, denoted as ID$-$ and C-OOD$-$. 
Within this MS-OOD framework, the objective of OOD detection methods is to accept correctly predicted examples and reject incorrectly predicted examples.

Figure~\ref{fig:MS-OOD-setting} presents the distribution of OOD scores measured on ID and \proposal (C-OOD) datasets under the MS-OOD framework. 
ViM~\cite{wang2022vim}, the current SOTA method for S-OOD detection, still exhibits a significant overlapping area between \textit{ImageNet-ES$+$} and \textit{ImageNet-ES$-$}. This confirms that methods developed to detect S-OOD cannot handle C-OOD data properly solely by modifying the underlying framework. 
On the other hand, MSP, an older method usually serving as a baseline, shows a clearer score separation between \textit{ImageNet-ES$+$} and \textit{ImageNet-ES$-$}.

\begin{figure}
\centering
    \includegraphics[width=.8\linewidth]{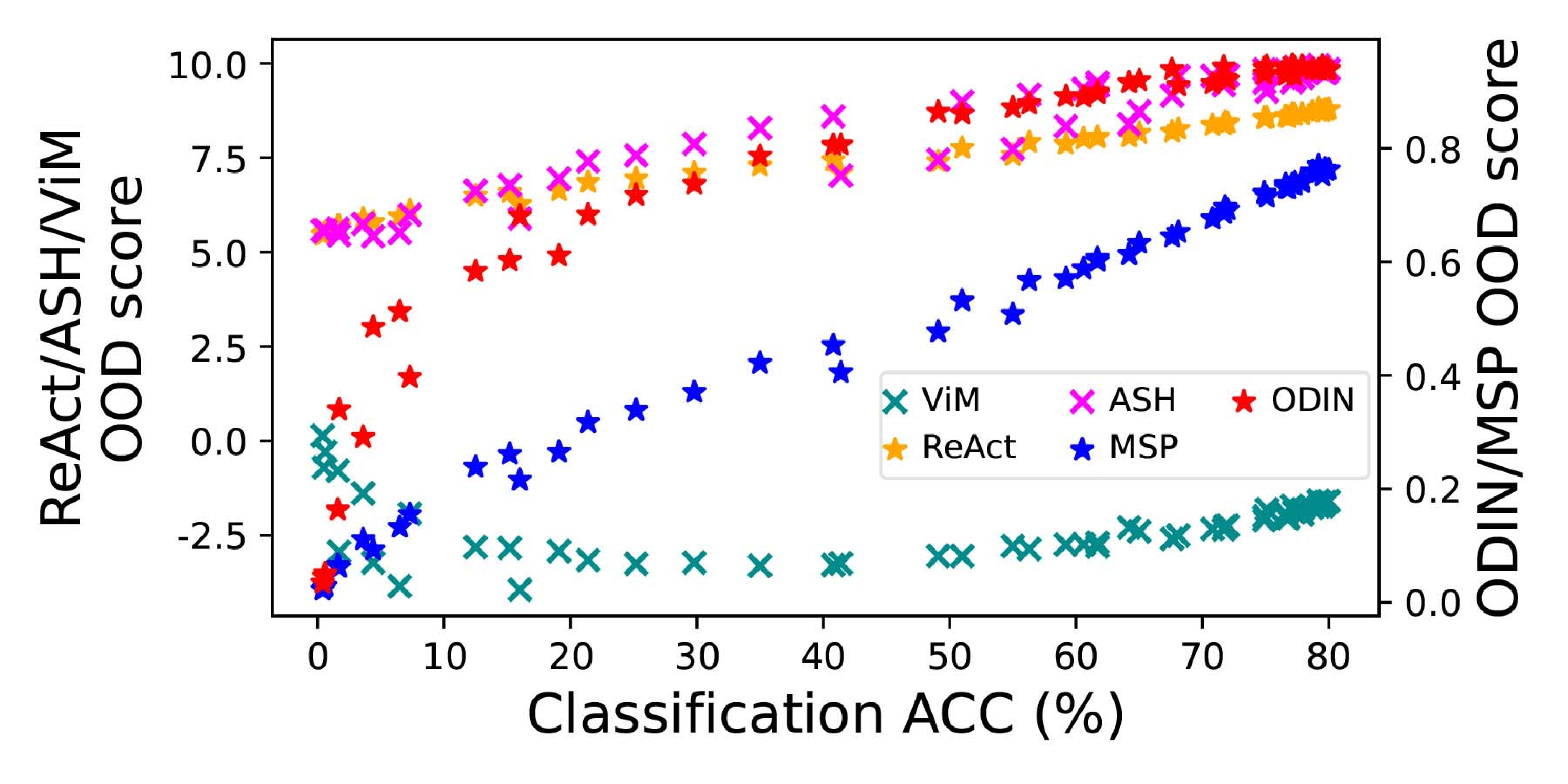}
    \vspace{-3ex}
    \caption{Each point represents the OOD score measured on the single parameter setting of \proposal.}
    \vspace{-3ex}
    \label{fig:acc-oodscore}
\end{figure}

\vspace{-2ex}
\subsubsection{Evaluation of OOD Detection Methods}
\vspace{-1ex}

Next, we evaluate 54 manual environmental/sensor variations in the \proposal test set in terms of classification accuracy and OOD scores. The OOD scores are obtained using five methods (MSP~\cite{hendrycks2016baseline}, ODIN~\cite{liang2017enhancing}, ReAct~\cite{react}, ASH~\cite{ash} and ViM~\cite{wang2022vim}) within the MS-OOD framework.

Figure~\ref{fig:acc-oodscore} showcases both accuracy and OOD scores for each setting, averaged over 1,000 samples out of 200 classes. Given that the MS-OOD framework defines OOD based on model prediction results, the OOD score is expected to increase with classification accuracy.
Our results reveal that the older methods, MSP~\cite{hendrycks2016baseline} and ODIN~\cite{liang2017enhancing}, provide relatively desirable correlation between accuracy and OOD score. In contrast, more recent methods (ASH~\cite{ash}, ReAct~\cite{react} and ViM~\cite{wang2022vim}) demonstrate a weaker relationship between accuracy and OOD score. Particularly, ViM shows a rapid increase in OOD scores as accuracy approaches zero; ViM tends to accept numerous samples as ID even when they are misclassified by the model.

\begin{figure}
    \begin{center}
        \begin{subfigure}{0.49\linewidth}
            \includegraphics[width=\linewidth]{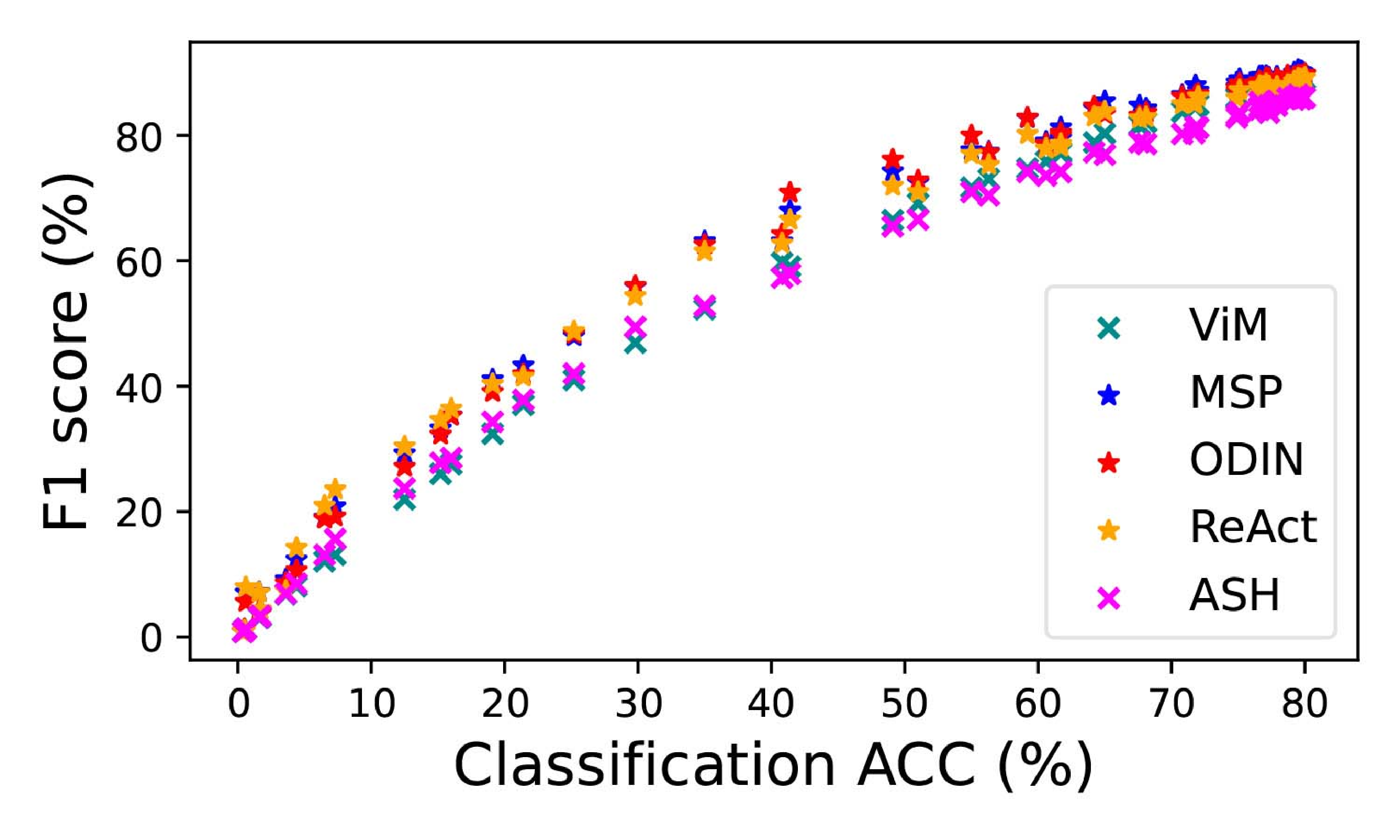}
            \vspace{-3ex}
            \caption{C-OOD (\proposal)}
            \label{fig:f1score}
        \end{subfigure}
        \begin{subfigure}{0.49\linewidth}
            \includegraphics[width=\linewidth]{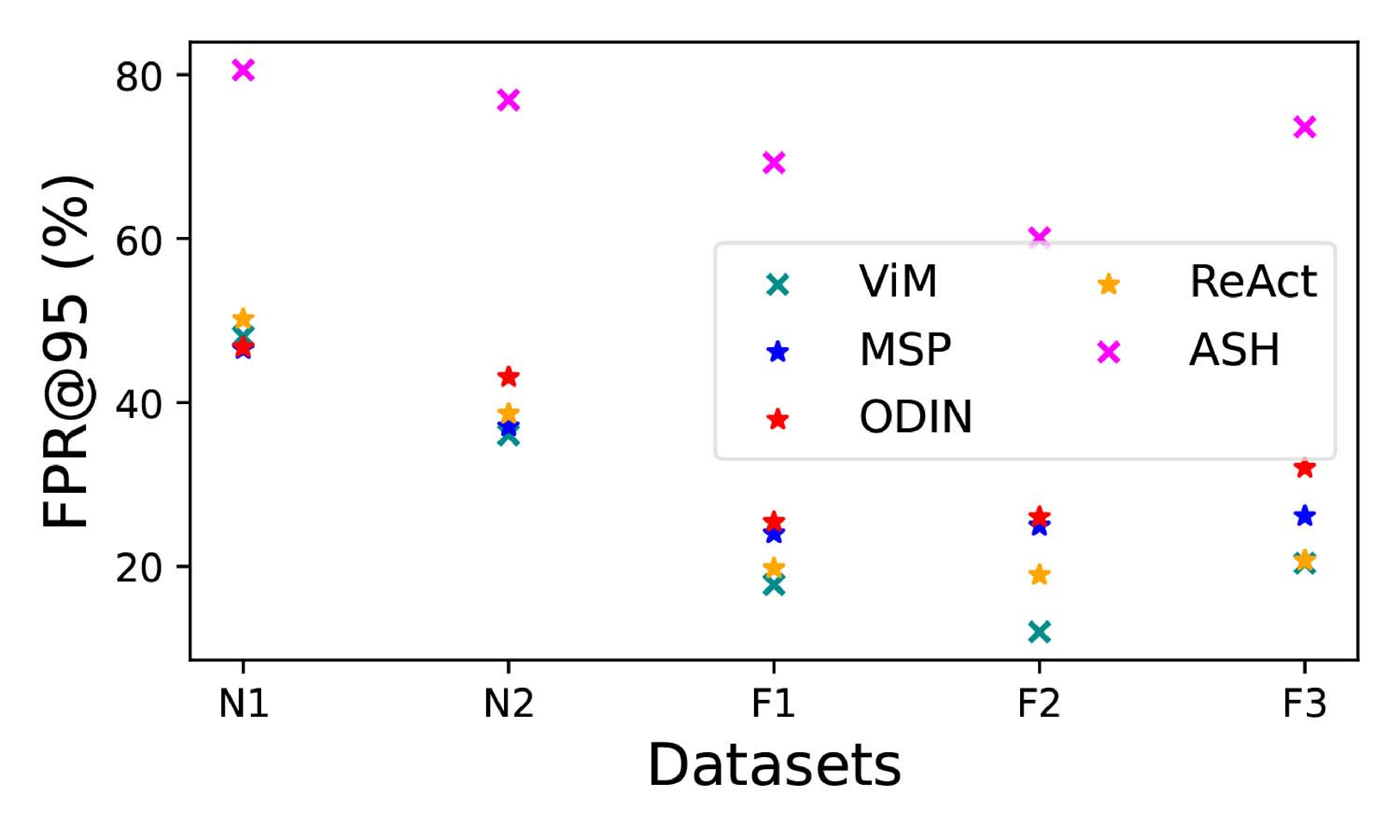}
            \vspace{-3ex}
            \caption{S-OOD}
            \label{fig:fpr}
        \end{subfigure}
        \vspace{-1ex}
    \caption{%\textcolor{blue}{[updated]} 
    Performance of OOD methods with C-OOD and S-OOD. (a) Each point is the F1 score measured on a parameter setting of \proposal. (b) N1: SSB-hard~\cite{SSB}, N2: NINCO~\cite{ninco}, F1: iNaturalist~\cite{Horn2018CVPR}, F2: Texture~\cite{Cimpoi2014CVPR}, F3: Openimage-O~\cite{wang2022vim}}
    \label{fig:ood-performance}
    \end{center}
    \vspace{-5ex}
\end{figure}

In addition, Figure~\ref{fig:f1score} shows the OOD detection performance 
for each environmental/sensor setting of the \proposal test set, in terms of F1 score used in~\cite{averly2023unified}. The results reveal that MSP and its advanced versions, ODIN and ReAct, consistently outperform ViM and ASH. Meanwhile the latest ViM and ASH emerge as the least effective among the five. The detection errors observed in ViM can be attributed to its inability to recognize unseen features in C-OOD. A more thorough explanation is in Appendix.

For further insight, Figure~\ref{fig:fpr} employs five S-OOD datasets as benchmarks, comprising two for near-OOD (SSB-hard~\cite{SSB}, NINCO~\cite{ninco}) and three for far-OOD (iNaturalist~\cite{Horn2018CVPR}, Texture~\cite{Cimpoi2014CVPR}, and Openimage-O~\cite{wang2022vim}). 
In contrast to the results in \proposal (Figure~\ref{fig:f1score}), 
Figure~\ref{fig:fpr} underscores that ViM provides superior performance compared to MSP, ODIN, ReAct and ASH, confirming the effectiveness of latest methods in detecting semantic shifts. 

Overall, our findings suggest that the evolution of OOD detection methods over recent years might have been biased towards S-OOD handling, potentially retrograding in terms of C-OOD handling. As of now, there is no single method that excels in both C-OOD and S-OOD detection. Given the importance of addressing covariate shifts in real-world applications, future research on OOD detection should integrate and advance both S-OOD and C-OOD aspects.

% \subsubsection{Robustness enhancements}
\subsection{Domain Generalization}

In this section, we investigate the impact of domain generalization techniques on enhancing the robustness in the environment and sensor domains. As a baseline, we finetune ResNet-50 using only the subset of ImageNet (IN) that precisely matches the images corresponding to the validation split from \proposal, incorporating composition-related augmentations such as crop, resize and flip.

For comparison schemes, we employ both basic and advanced digital augmentations. Basic augmentations include color-jitter, solarize and posterize, while advanced augmentations include DeepAugment~\cite{hendrycks2021many} and AugMix~\cite{hendrycks2020augmix}.
To explore the potential of finetuning with our real-world perturbations,  %seeking to ascertain whether such an approach contributes to the improved robustness of the model in environment and sensor domain. 
we replace half of the finetuning images with randomly sampled images from the validation set of \proposal. We exclude some images that are too far distorted to be identifiable, utilizing an image similarity metric LPIPS~\cite{zhang2018perceptual}.
We evaluate each finetuning result on the test sets of IN, IN-C~\cite{hendrycks2019benchmarking} and \proposal. Since \proposal contains only a subset of images from 200 classes from IN, we use the same subset of IN and IN-C corresponding to the test set of \proposal for fair comparison.

\begin{table}[t]
\centering
  \caption{Evaluation with different robustness enhancing strategies. The result is based on ResNet-50. (IN: ImageNet)}
  \vspace{-2ex}
  \label{tab:DG_enhance}

\resizebox{\linewidth}{!}{ %< auto-adjusts font size to fill line
  \begin{tabular}{cccccccc}
    \toprule
    \multirow{2}{*}{ID}         & \multirow{2}{*}{Comp.aug}             & Basic                  & Advanced    & Incl. & \multicolumn{3}{c}{Eval dataset} \\
     & & digital aug & digital aug & \proposal & IN & IN-C &  \proposal  \\
    \midrule
    1 & \checkmark & & &                                  & 85.8 & 51.0 & 49.6 \\
    2 & \checkmark & \checkmark & &                       & 85.8 & 51.7 & 50.4 \\
    3 & \checkmark & \checkmark & \checkmark &            & 85.5 & 57.4 & 49.1 \\
    4 & \checkmark & &                       & \checkmark & \textbf{86.0} & 51.8 & \textbf{55.8} \\
    5 & \checkmark & \checkmark &            & \checkmark & 85.8 & 51.4 & 54.5 \\
    6 & \checkmark & \checkmark & \checkmark & \checkmark & 84.0 & \textbf{57.9} & 53.7 \\
  \bottomrule

    \end{tabular}
} % \resizebox
\vspace{-3ex}
\end{table}

The evaluation results are summarized in Table~\ref{tab:DG_enhance}. Experiment 2 shows that basic augmentations lead to improved accuracy in both IN-C and \proposal. 
However, our findings in experiment 3 contradict prior work~\cite{hendrycks2021many,hendrycks2020augmix}. While more aggressive augmentations, such as AugMix and DeepAugment, significantly improve robustness on IN-C, these strategies result in performance degradation when predicting images with our real-world perturbations.

Experiments 4 to 6 evaluate the impact of learning augmentations in environmental and sensor domains in addition to conventional digital augmentations. 
Learning these real-world perturbations consistently  improves robustness in \proposal, demonstrating its effectiveness in real-world applications. Furthermore, experiments 4 and 6 show that adopting environmental/sensor augmentations further improves robustness on the conventional benchmark IN-C.

In summary, these results verify that augmentations in the environmental and sensor domains can provide additional valuable information absent in conventional augmentation schemes. The impact becomes more significant in real-world applications involving cameras.

%%%%%%%%%%%%%%%%%%%%%%%%%%%%%%%%%%%%%%%
%%%%%%%%%%%%%%%%%%%%%%%%%%%%%%%%%%%%%%%
\subsection{Sensor Parameter Control}

To investigate the impact of sensor parameter control on model performance, we evaluate the performance of three different subsets of \proposal test split. \textbf{Auto exposure} includes 10,000 samples captured with the default auto exposure (AE) setting. \textbf{All params} includes 54,000 samples captured with 27 different manual parameters settings. \textbf{Best} includes 2,000 samples captured with the manual parameter setting that provides the best accuracy among the test split.

We employ several models for generalizability. The baseline is ResNet-50~\cite{he2016deep}, trained with a vanilla training scheme on ImageNet (IN)-1K. To explore whether well-configured sensor parameters could enhance the model performance to the level of  domain-generalizated (DG) models, we also evaluate ResNet-50 trained on IN-21K with DeepAugment~\cite{hendrycks2021many} and AugMix~\cite{hendrycks2020augmix}. In addition, we examine whether a larger model demonstrates more robustness by evaluating ResNet-152.

Furthermore, we employ EfficientNet-B0/B3~\cite{pmlr-v97-tan19a} to test the lightweight model architecture's validity on \proposal. SwinV2-T/B~\cite{liu2022swin} are chosen as representative models from transformer-based architecture, known for its robustness~\cite{bhojanapalli2021understanding, hendrycks2019benchmarking}. 
% We test both tiny and base models, denoted as SwinV2-T and SwinV2-B, respectively, to analyze the influence of model size.
OpenCLIP-b/h~\cite{ilharco} and DINOv2-b/g~\cite{oquab2023dinov2} are selected as domain-generalized versions of SwinV2. 
All model weights are obtained from PyTorch~\cite{paszke2019pytorch}, except for the DG version of ResNet-50, which is released in~\cite{hendrycks2021many}.
\vspace{-2ex}

\begin{table}[t]
\centering
  \caption{Evaluation of various models on \proposal. (IN: ImageNet, AE: Auto exposure)}
  \label{tab:param_control}
  \vspace{-2ex}

\resizebox{\linewidth}{!}{ %< auto-adjusts font size to fill line
  \begin{tabular}{cccccccc}
    \toprule

    \multirow{2}{*}{Model} & Num.  & Pretraining & \multirow{2}{*}{DG method}  & \multirow{2}{*}{IN}   & \multicolumn{3}{c}{\proposal} \\
                           & Params & Dataset     &                             &                       & AE & All params & Best \\
    \midrule
    
    \multirow{3}{*}{ResNet-50~\cite{he2016deep}}  &  \multirow{3}{*}{26M} & IN-1K  & -                    & 86.3 & 32.2 & 50.2 & 80.1 \\
        &                       & \multirow{2}{*}{IN-21K} & DeepAugment~\cite{hendrycks2021many} 
                                  & \multirow{2}{*}{87.0} & \multirow{2}{*}{53.3} & \multirow{2}{*}{61.4} & \multirow{2}{*}{84.0} \\
            &                   &   & +AugMix~\cite{hendrycks2020augmix}&&&&\\
    ResNet-152~\cite{he2016deep}                  & 60M & IN-1K & -                                       & 87.6 & 41.1 & 54.3 & 83.3 \\
    \hline
    Efficientnet-B0~\cite{pmlr-v97-tan19a}              & 5M & IN-1K & -     & 88.1 & 51.4 & 58.1 & 83.8 \\    
    Efficientnet-B3~\cite{pmlr-v97-tan19a}              & 12M & IN-1K & -     & 88.3 & 62.0 & 66.2 & 86.8 \\ 
    \hline
    SwinV2-T~\cite{liu2022swin}                    & 28M & IN-1K & -                                       & 90.7 & 54.2 & 63.1 & 86.8 \\
    
    SwinV2-B~\cite{liu2022swin}                    & 88M & IN-1K & -                                       & 92.0 & 60.1 & 65.6 & 89.0 \\
    \hline
    OpenCLIP-b~\cite{ilharco}              & 87M & LAION-2B & Text-guided pretrain     & 94.3 & 66.3 & 71.0 & 92.7 \\    
    OpenCLIP-h~\cite{ilharco}              & 632M & LAION-2B & Text-guided pretrain     & 94.7 & 79.1 & 77.6 & 94.7 \\    
    \hline
    DINOv2-b~\cite{oquab2023dinov2}              & 90M & LVD-142M & Dataset curation     & 93.6 & 74.5 & 73.9 & 92.2 \\    
    DINOv2-g~\cite{oquab2023dinov2}              & 1.1B & LVD-142M & Dataset curation     & 94.7 & 84.3 & 79.6 & 94.2 \\    
  \bottomrule
    \end{tabular}
} % \resizebox
\vspace{-3ex}
\end{table}

%%%%%%%%%%%%%%%%%%%%%%%%%%%%%%%%%
%%%%%%%%%%%%%%%%%%%%%%%%%%%%%%%%%
\subsubsection{Potential of Sensor Control}
\vspace{-0.5ex}
Table~\ref{tab:param_control} summarizes the results. 
Firstly, DG techniques and pretraining on larger datasets consistently improve robustness on \proposal. For instance, in the All params case, DG version of ResNet-50 improves the test accuracy to plain ResNet-50 by 11.2. 

In addition, sensor parameter tuning turns out to be as important as domain generalization and model size. 
First, the auto exposure option degrades performance of all models compared to the accuracy on original images (IN); %. This demonstrates that %the impact of sensor parameters is critical and 
the current auto exposure does not provide optimal parameters for model predictions. 
Conversely, the Best case reveals that with well-tuned sensor parameters, performance can be improved remarkably. 
The Best case improves prediction accuracy over the Auto exposure case by \textbf{9.9$\sim$47.9} and the All params case by \textbf{14.6$\sim$29.9}. 
Surprisingly, EfficientNet-B0 with the Best even outperforms OpenCLIP-h in the Auto exposure and All params cases, despite OpenCLIP-h having around 120$\times$ more parameters, learning from 160$\times$ more training data and employing domain generalization.

These findings suggest that research might have over-emphasized model improvement, possibly overlooking the importance of proper input data generation. However, mitigating distribution shifts through sensor control proves to be valuable regardless of model size and DG techniques. The performance gain from sensor control can even surpass that achieved through larger model size, more training data, advanced architectures, and DG techniques.  
If designed efficiently, sensor control can be an enabling factor of mobile applications where computational resources are limited.

\begin{table}[t]
\centering
  \caption{Impact of environmental factor. The difference is calculated between the accuracy measured when light is on and off. We provide the difference for auto exposure setting and the maximum of difference amongst all manual parameter settings. More details could be found in Appendix.} %\section{appendix}}
  \label{tab:param_light}
  \vspace{-2ex}

\resizebox{\linewidth}{!}{ %< auto-adjusts font size to fill line
  \begin{tabular}{ccccc}
    \toprule
    Model               & ResNet-50 & ResNet-152 & SwinV2-B & DINOv2 \\
    \midrule
    Auto exposure       & 4.4  & 3.0  & 4.3  & 2.4 \\
    Max. of all params  & 11.7 & 14.9 & 15.7 & 18.2 \\
    
  \bottomrule

    \end{tabular}
} % \resizebox
\vspace{-2ex}
\end{table}

\begin{figure}
    \centering
    \includegraphics[width=\linewidth]{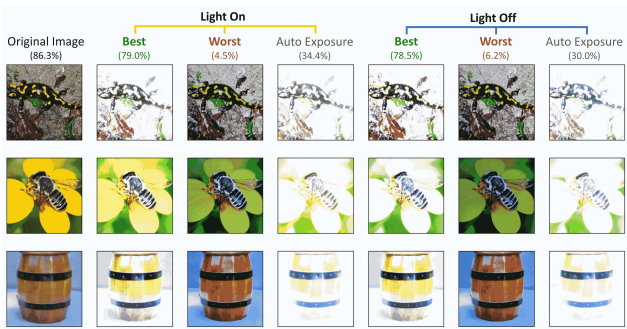}    
    \vspace{-4ex} 
\caption{Qualitative results of \proposal: AE: Auto Exposure, Best/Worst: Sampled images from the parameter setting whose accuracy is highest 5 or lowest 5. Accuracy on ResNet-50 is also presented for each group.}
    \label{fig:paramtercontrol-quality}
\vspace{-3ex}    
\end{figure}

%%%%%%%%%%%%%%%%%%%%%%%%%%%%%%%%%
%%%%%%%%%%%%%%%%%%%%%%%%%%%%%%%%%
\vspace{-1ex}
\subsubsection{Direction of Sensor Control}
\vspace{-0.5ex}

Table~\ref{tab:param_light} summarizes the impact of lighting conditions (on and off) on model accuracy. Specifically, the table shows the difference in the model accuracy caused by changes in lighting when using the same sensor parameter option. 
Notably, the auto-exposure option fails to ensure consistent model performance, showing accuracy differences of 2.4$\sim$4.4 across diverse models. Moreover, manual configuration of sensor parameters is  also susceptible to the impact of environmental variables, resulting in an accuracy variance of up to 18.2. 
Our results pose a challenge for future research on sensor control. It should focus on adaptively controlling parameters based on given environments, instead of attempting to find universally optimal parameters.

To obtain further insights into sensor control, we perform a qualitative analysis on \proposal, as described in Figure \ref{fig:paramtercontrol-quality}.
%
%we have gained valuable insights into the camera sensor parameter control method for restoring the original performance of models. 
Interestingly, these examples show that images visually appealing to humans do not necessarily yield the best prediction outcomes. Images captured with the Best options sometimes appear significantly lighter, posing a challenge for human observers to discern the underlying semantic information. Conversely, some images captured with the worst-performing options are more conducive to semantic interpretation by humans. Lastly, images captured by  controlled parameters are occasionally predicted more accurately than the original samples. 
These results imply that sensor control research should prioritize features that the model can leverage effectively, rather than relying solely on human intuition. A more detailed examination of the camera parameters is in Appendix.

\section{Conclusion and Future Work}

In this study, we investigated distribution shifts resulting from perturbations in both environmental and sensor domains. To achieve this goal, we established a controllable testbed, named \testbed, for image acquisition across diverse environmental and camera sensor configurations. We curated a new dataset of 202k images, called \proposal.

Employing \proposal, we have conducted comprehensive studies in OOD detection, domain generalization and camera sensor control. 
With respect to OOD detection, our findings indicated limitations in the widely used semantics-centric framework. We proposed that OOD detection should extend its scope to incorporate both S-OOD and C-OOD.
Regarding domain generalization, we demonstrated that environmental and sensor-related augmentations offer  additional useful information to the model, improving robustness in both conventional and \proposal benchmarks. 
Finally, we discovered that well-tuned sensor parameters can enable a lightweight, basic model to perform comparably to or better than more advanced models. Sensor control necessitates a model-centric design instead of relying solely on human aesthetics.
We hope that these insights will inspire future research and \proposal will be utilized in addressing real-world distribution shifts.

\vspace{1ex}\noindent
\textbf{Limitations and Future Work.}
Taking photos of displays in \testbed is reproducible, controllable, and scalable, but it might not fully consider the interaction between real 3D objects and light sources, and the non-luminous properties of real objects. It would be more realistic to replace displays with real objects or printed photos. Sensor control can support applications like autonomous driving and surveillance cameras, which require image capture in various environments. But since it primarily handles physical light, it needs to be combined with digital post-processing. For practical training of a neural net for sensor control, gradients need to be computed without extra photos.

\section*{Acknowledgements}
\vspace{-0.5ex}
This research was supported in part by  Institute
of Information \& communications Technology Planning \& Evaluation (IITP) grant funded by the Korea government (MSIT) (No. RS-2023-00223530), and in part by the National Research Foundation (NRF) of Korea grants funded by the
Korea government (MSIT) (No. RS-2023-00212780, No. RS-2023-00222663 and No. RS-2023-00265147). Hyung-Sin Kim is the corresponding author.

\newpage
{
    \small
    \bibliographystyle{ieeenat_fullname}
    \bibliography{main}

\begin{thebibliography}{44}
\providecommand{\natexlab}[1]{#1}
\providecommand{\url}[1]{\texttt{#1}}
\expandafter\ifx\csname urlstyle\endcsname\relax
  \providecommand{\doi}[1]{doi: #1}\else
  \providecommand{\doi}{doi: \begingroup \urlstyle{rm}\Url}\fi

\bibitem[Averly and Chao(2023)]{averly2023unified}
Reza Averly and Wei-Lun Chao.
\newblock Unified out-of-distribution detection: A model-specific perspective.
\newblock \emph{International Conference on Computer Vision (ICCV)}, 2023.

\bibitem[Bhojanapalli et~al.(2021)Bhojanapalli, Chakrabarti, Glasner, Li, Unterthiner, and Veit]{bhojanapalli2021understanding}
Srinadh Bhojanapalli, Ayan Chakrabarti, Daniel Glasner, Daliang Li, Thomas Unterthiner, and Andreas Veit.
\newblock Understanding robustness of transformers for image classification.
\newblock In \emph{Proceedings of the IEEE/CVF international conference on computer vision}, pages 10231--10241, 2021.

\bibitem[Bitterwolf et~al.(2023)Bitterwolf, Mueller, and Hein]{ninco}
Julian Bitterwolf, Maximilian Mueller, and Matthias Hein.
\newblock In or out? fixing imagenet out-of-distribution detection evaluation.
\newblock In \emph{ICML}, 2023.

\bibitem[Cimpoi et~al.(2014)Cimpoi, Maji, Kokkinos, Mohamed, and Vedaldi]{Cimpoi2014CVPR}
Mircea Cimpoi, Subhransu Maji, Iasonas Kokkinos, Sammy Mohamed, and Andrea Vedaldi.
\newblock Describing textures in the wild.
\newblock In \emph{Proceedings of the IEEE Conference on Computer Vision and Pattern Recognition (CVPR)}, 2014.

\bibitem[Cubuk et~al.(2019)Cubuk, Zoph, Mane, Vasudevan, and Le]{Cubuk}
Ekin~D. Cubuk, Barret Zoph, Dandelion Mane, Vijay Vasudevan, and Quoc~V. Le.
\newblock Autoaugment: Learning augmentation strategies from data.
\newblock In \emph{Proceedings of the IEEE/CVF Conference on Computer Vision and Pattern Recognition (CVPR)}, 2019.

\bibitem[Djurisic et~al.(2023)Djurisic, Bozanic, Ashok, and Liu]{ash}
Andrija Djurisic, Nebojsa Bozanic, Arjun Ashok, and Rosanne Liu.
\newblock Extremely simple activation shaping for out-of-distribution detection.
\newblock \emph{The Eleventh International Conference on Learning Representations}, 2023.

\bibitem[Goodfellow et~al.(2014)Goodfellow, Shlens, and Szegedy]{Goodfellow2014ExplainingAH}
Ian~J. Goodfellow, Jonathon Shlens, and Christian Szegedy.
\newblock Explaining and harnessing adversarial examples.
\newblock \emph{CoRR}, abs/1412.6572, 2014.

\bibitem[He et~al.(2016)He, Zhang, Ren, and Sun]{he2016deep}
Kaiming He, Xiangyu Zhang, Shaoqing Ren, and Jian Sun.
\newblock Deep residual learning for image recognition.
\newblock In \emph{Proceedings of the IEEE conference on computer vision and pattern recognition}, pages 770--778, 2016.

\bibitem[He et~al.(2022)He, Chen, Xie, Li, Doll{\'a}r, and Girshick]{he2022masked}
Kaiming He, Xinlei Chen, Saining Xie, Yanghao Li, Piotr Doll{\'a}r, and Ross Girshick.
\newblock Masked autoencoders are scalable vision learners.
\newblock In \emph{Proceedings of the IEEE/CVF conference on computer vision and pattern recognition}, pages 16000--16009, 2022.

\bibitem[Hendrycks and Dietterich(2019)]{hendrycks2019benchmarking}
Dan Hendrycks and Thomas Dietterich.
\newblock Benchmarking neural network robustness to common corruptions and perturbations.
\newblock \emph{arXiv e-prints}, pages arXiv--1903, 2019.

\bibitem[Hendrycks and Gimpel(2017)]{hendrycks2016baseline}
Dan Hendrycks and Kevin Gimpel.
\newblock A baseline for detecting misclassified and out-of-distribution examples in neural networks.
\newblock \emph{International Conference on Learning Representations}, 2017.

\bibitem[Hendrycks et~al.(2020)Hendrycks, Mu, Cubuk, Zoph, Gilmer, and Lakshminarayanan]{hendrycks2020augmix}
Dan Hendrycks, Norman Mu, Ekin~D. Cubuk, Barret Zoph, Justin Gilmer, and Balaji Lakshminarayanan.
\newblock Augmix: A simple data processing method to improve robustness and uncertainty.
\newblock \emph{Proceedings of the International Conference on Learning Representations (ICLR)}, 2020.

\bibitem[Hendrycks et~al.(2021{\natexlab{a}})Hendrycks, Basart, Mu, Kadavath, Wang, Dorundo, Desai, Zhu, Parajuli, Guo, et~al.]{hendrycks2021many}
Dan Hendrycks, Steven Basart, Norman Mu, Saurav Kadavath, Frank Wang, Evan Dorundo, Rahul Desai, Tyler Zhu, Samyak Parajuli, Mike Guo, et~al.
\newblock The many faces of robustness: A critical analysis of out-of-distribution generalization.
\newblock In \emph{Proceedings of the IEEE/CVF International Conference on Computer Vision}, pages 8340--8349, 2021{\natexlab{a}}.

\bibitem[Hendrycks et~al.(2021{\natexlab{b}})Hendrycks, Zhao, Basart, Steinhardt, and Song]{hendrycks2021natural}
Dan Hendrycks, Kevin Zhao, Steven Basart, Jacob Steinhardt, and Dawn Song.
\newblock Natural adversarial examples.
\newblock In \emph{Proceedings of the IEEE/CVF Conference on Computer Vision and Pattern Recognition}, pages 15262--15271, 2021{\natexlab{b}}.

\bibitem[Hendrycks et~al.(2022)Hendrycks, Basart, Mazeika, Zou, Kwon, Mostajabi, Steinhardt, and Song]{hendrycks2019scaling}
Dan Hendrycks, Steven Basart, Mantas Mazeika, Andy Zou, Joseph Kwon, Mohammadreza Mostajabi, Jacob Steinhardt, and Dawn Song.
\newblock Scaling out-of-distribution detection for real-world settings.
\newblock \emph{Proceedings of the 39th International Conference on Machine Learning}, pages 8759--8773, 2022.

\bibitem[Hsu et~al.(2020)Hsu, Shen, Jin, and Kira]{hsu2020generalized}
Yen-Chang Hsu, Yilin Shen, Hongxia Jin, and Zsolt Kira.
\newblock Generalized odin: Detecting out-of-distribution image without learning from out-of-distribution data.
\newblock In \emph{Proceedings of the IEEE/CVF Conference on Computer Vision and Pattern Recognition}, pages 10951--10960, 2020.

\bibitem[Ilharco et~al.(2021)Ilharco, Wortsman, Wightman, Gordon, Carlini, Taori, Dave, Shankar, Namkoong, Miller, Hajishirzi, Farhadi, and Schmidt]{ilharco}
Gabriel Ilharco, Mitchell Wortsman, Ross Wightman, Cade Gordon, Nicholas Carlini, Rohan Taori, Achal Dave, Vaishaal Shankar, Hongseok Namkoong, John Miller, Hannaneh Hajishirzi, Ali Farhadi, and Ludwig Schmidt.
\newblock Openclip, 2021.

\bibitem[Jishnu~Mukhoti et~al.(2023)Jishnu~Mukhoti, Bor-Chun~Chen, Philip H.S.~Torr, and Lim]{raisingbar2023}
Tsung-Yu~Lin Jishnu~Mukhoti, Ashish~Shah Bor-Chun~Chen, Puneet K.~Dokania Philip H.S.~Torr, and Ser-Nam Lim.
\newblock Raising the bar on the evaluation of out-of-distribution detection.
\newblock In \emph{Proceedings of the IEEE/CVF International Conference on Computer Vision (ICCV) Workshops}, pages 4365--4375, 2023.

\bibitem[Lee et~al.(2018)Lee, Lee, Lee, and Shin]{lee2018simple}
Kimin Lee, Kibok Lee, Honglak Lee, and Jinwoo Shin.
\newblock A simple unified framework for detecting out-of-distribution samples and adversarial attacks.
\newblock \emph{Advances in neural information processing systems}, 31, 2018.

\bibitem[Li et~al.(2023)Li, Chen, Zhu, Wang, Zhang, and Xue]{li2023imagenet}
Xiaodan Li, Yuefeng Chen, Yao Zhu, Shuhui Wang, Rong Zhang, and Hui Xue.
\newblock Imagenet-e: Benchmarking neural network robustness via attribute editing.
\newblock In \emph{Proceedings of the IEEE/CVF Conference on Computer Vision and Pattern Recognition}, pages 20371--20381, 2023.

\bibitem[Liang et~al.(2018)Liang, Li, and Srikant]{liang2017enhancing}
Shiyu Liang, Yixuan Li, and R. Srikant.
\newblock Enhancing the reliability of out-of-distribution image detection in neural networks.
\newblock \emph{International Conference on Learning Representations}, 2018.

\bibitem[Liu et~al.(2020)Liu, Wang, Owens, and Li]{liu2020energy}
Weitang Liu, Xiaoyun Wang, John Owens, and Yixuan Li.
\newblock Energy-based out-of-distribution detection.
\newblock \emph{Advances in neural information processing systems}, 33:\penalty0 21464--21475, 2020.

\bibitem[Liu et~al.(2022{\natexlab{a}})Liu, Hu, Lin, Yao, Xie, Wei, Ning, Cao, Zhang, Dong, et~al.]{liu2022swin}
Ze Liu, Han Hu, Yutong Lin, Zhuliang Yao, Zhenda Xie, Yixuan Wei, Jia Ning, Yue Cao, Zheng Zhang, Li Dong, et~al.
\newblock Swin transformer v2: Scaling up capacity and resolution.
\newblock In \emph{Proceedings of the IEEE/CVF conference on computer vision and pattern recognition}, pages 12009--12019, 2022{\natexlab{a}}.

\bibitem[Liu et~al.(2022{\natexlab{b}})Liu, Mao, Wu, Feichtenhofer, Darrell, and Xie]{liu2022convnet}
Zhuang Liu, Hanzi Mao, Chao-Yuan Wu, Christoph Feichtenhofer, Trevor Darrell, and Saining Xie.
\newblock A convnet for the 2020s.
\newblock In \emph{Proceedings of the IEEE/CVF conference on computer vision and pattern recognition}, pages 11976--11986, 2022{\natexlab{b}}.

\bibitem[Madry et~al.(2018)Madry, Makelov, Schmidt, Tsipras, and Vladu]{madry2018towards}
Aleksander Madry, Aleksandar Makelov, Ludwig Schmidt, Dimitris Tsipras, and Adrian Vladu.
\newblock Towards deep learning models resistant to adversarial attacks.
\newblock In \emph{International Conference on Learning Representations}, 2018.

\bibitem[Oquab et~al.(2023)Oquab, Darcet, Moutakanni, Vo, Szafraniec, Khalidov, Fernandez, Haziza, Massa, El-Nouby, et~al.]{oquab2023dinov2}
Maxime Oquab, Timoth{\'e}e Darcet, Th{\'e}o Moutakanni, Huy Vo, Marc Szafraniec, Vasil Khalidov, Pierre Fernandez, Daniel Haziza, Francisco Massa, Alaaeldin El-Nouby, et~al.
\newblock Dinov2: Learning robust visual features without supervision.
\newblock \emph{arXiv preprint arXiv:2304.07193}, 2023.

\bibitem[Paszke et~al.(2019)Paszke, Gross, Massa, Lerer, Bradbury, Chanan, Killeen, Lin, Gimelshein, Antiga, et~al.]{paszke2019pytorch}
Adam Paszke, Sam Gross, Francisco Massa, Adam Lerer, James Bradbury, Gregory Chanan, Trevor Killeen, Zeming Lin, Natalia Gimelshein, Luca Antiga, et~al.
\newblock Pytorch: An imperative style, high-performance deep learning library.
\newblock \emph{Advances in neural information processing systems}, 32, 2019.

\bibitem[Qizhou~Wang et~al.(2022)Qizhou~Wang, Yonggang~Zhang, Chen~Gong, and Han]{watermarking2022}
Feng~Liu Qizhou~Wang, Jing~Zhang Yonggang~Zhang, Tongliang~Liu Chen~Gong, and Bo Han.
\newblock Watermarking for out-of-distribution detection.
\newblock In \emph{Advances in Neural Information Processing Systems 35 (NeurIPS)}, 2022.

\bibitem[Rusak et~al.(2022)Rusak, Schneider, Gehler, Bringmann, Brendel, and Bethge]{rusak2022imagenet}
Evgenia Rusak, Steffen Schneider, Peter~Vincent Gehler, Oliver Bringmann, Wieland Brendel, and Matthias Bethge.
\newblock Imagenet-d: A new challenging robustness dataset inspired by domain adaptation.
\newblock In \emph{ICML 2022 Shift Happens Workshop}, 2022.

\bibitem[Sun et~al.(2021)Sun, Guo, and Li]{react}
Yiyou Sun, Chuan Guo, and Yixuan Li.
\newblock React: Out-of-distribution detection with rectified activations.
\newblock \emph{Advances in Neural Information Processing Systems}, 34:\penalty0 144--157, 2021.

\bibitem[Sun et~al.(2022)Sun, Ming, Zhu, and Li]{sun2022out}
Yiyou Sun, Yifei Ming, Xiaojin Zhu, and Yixuan Li.
\newblock Out-of-distribution detection with deep nearest neighbors.
\newblock In \emph{International Conference on Machine Learning}, pages 20827--20840. PMLR, 2022.

\bibitem[Tan and Le(2019)]{pmlr-v97-tan19a}
Mingxing Tan and Quoc Le.
\newblock {E}fficient{N}et: Rethinking model scaling for convolutional neural networks.
\newblock In \emph{Proceedings of the 36th International Conference on Machine Learning}, pages 6105--6114. PMLR, 2019.

\bibitem[Tian et~al.(2021)Tian, Hsu, Shen, Jin, and Kira]{tian2021exploring}
Junjiao Tian, Yen-Chang Hsu, Yilin Shen, Hongxia Jin, and Zsolt Kira.
\newblock Exploring covariate and concept shift for out-of-distribution detection.
\newblock In \emph{NeurIPS 2021 Workshop on Distribution Shifts: Connecting Methods and Applications}, 2021.

\bibitem[Van~Horn et~al.(2018)Van~Horn, Mac~Aodha, Song, Cui, Sun, Shepard, Adam, Perona, and Belongie]{Horn2018CVPR}
Grant Van~Horn, Oisin Mac~Aodha, Yang Song, Yin Cui, Chen Sun, Alex Shepard, Hartwig Adam, Pietro Perona, and Serge Belongie.
\newblock The inaturalist species classification and detection dataset.
\newblock In \emph{Proceedings of the IEEE Conference on Computer Vision and Pattern Recognition (CVPR)}, 2018.

\bibitem[Vaze et~al.(2022)Vaze, Han, Vedaldi, and Zisserman]{SSB}
Sagar Vaze, Kai Han, Andrea Vedaldi, and Andrew Zisserman.
\newblock Open-set recognition: A good closed-set classifier is all you need.
\newblock In \emph{International Conference on Learning Representations}, 2022.

\bibitem[Wang et~al.(2019)Wang, Ge, Lipton, and Xing]{wang2019learning}
Haohan Wang, Songwei Ge, Zachary Lipton, and Eric~P Xing.
\newblock Learning robust global representations by penalizing local predictive power.
\newblock In \emph{Advances in Neural Information Processing Systems}, pages 10506--10518, 2019.

\bibitem[Wang et~al.(2022)Wang, Li, Feng, and Zhang]{wang2022vim}
Haoqi Wang, Zhizhong Li, Litong Feng, and Wayne Zhang.
\newblock Vim: Out-of-distribution with virtual-logit matching supplementary material.
\newblock In \emph{Proceedings of the IEEE/CVF conference on computer vision and pattern recognition}, pages 4921--4930, 2022.

\bibitem[Wu et~al.(2017)Wu, Zhang, and Xu]{wu2017tiny}
Jiayu Wu, Qixiang Zhang, and Guoxi Xu.
\newblock Tiny imagenet challenge.
\newblock \emph{Technical report}, 2017.

\bibitem[Yang et~al.(2021)Yang, Wang, Feng, Yan, Zheng, Zhang, and Liu]{yang2021semantically}
Jingkang Yang, Haoqi Wang, Litong Feng, Xiaopeng Yan, Huabin Zheng, Wayne Zhang, and Ziwei Liu.
\newblock Semantically coherent out-of-distribution detection.
\newblock In \emph{Proceedings of the IEEE/CVF International Conference on Computer Vision}, pages 8301--8309, 2021.

\bibitem[Yang et~al.(2023)Yang, Zhou, and Liu]{yang2023full}
Jingkang Yang, Kaiyang Zhou, and Ziwei Liu.
\newblock Full-spectrum out-of-distribution detection.
\newblock \emph{International Journal of Computer Vision}, pages 1--16, 2023.

\bibitem[Yun et~al.(2019)Yun, Han, Oh, Chun, Choe, and Yoo]{yun2019cutmix}
Sangdoo Yun, Dongyoon Han, Seong~Joon Oh, Sanghyuk Chun, Junsuk Choe, and Youngjoon Yoo.
\newblock Cutmix: Regularization strategy to train strong classifiers with localizable features.
\newblock In \emph{Proceedings of the IEEE/CVF international conference on computer vision}, pages 6023--6032, 2019.

\bibitem[Zhang et~al.(2023)Zhang, Yang, Wang, Wang, Lin, Zhang, Sun, Du, Zhou, Zhang, et~al.]{zhang2023openood}
Jingyang Zhang, Jingkang Yang, Pengyun Wang, Haoqi Wang, Yueqian Lin, Haoran Zhang, Yiyou Sun, Xuefeng Du, Kaiyang Zhou, Wayne Zhang, et~al.
\newblock Openood v1. 5: Enhanced benchmark for out-of-distribution detection.
\newblock \emph{arXiv preprint arXiv:2306.09301}, 2023.

\bibitem[Zhang et~al.(2018)Zhang, Isola, Efros, Shechtman, and Wang]{zhang2018perceptual}
Richard Zhang, Phillip Isola, Alexei~A Efros, Eli Shechtman, and Oliver Wang.
\newblock The unreasonable effectiveness of deep features as a perceptual metric.
\newblock In \emph{Proceedings of the IEEE conference on computer vision and pattern recognition}, 2018.

\bibitem[Zhou et~al.(2022)Zhou, Wei, Wang, Shen, Xie, Yuille, and Kong]{zhou2021ibot}
Jinghao Zhou, Chen Wei, Huiyu Wang, Wei Shen, Cihang Xie, Alan Yuille, and Tao Kong.
\newblock ibot: Image bert pre-training with online tokenizer.
\newblock \emph{International Conference on Learning Representations (ICLR)}, 2022.

\end{thebibliography}
}

\end{document}